\documentclass[acmsmall]{acmart}

\AtBeginDocument{%
  \providecommand\BibTeX{{%
    \normalfont B\kern-0.5em{\scshape i\kern-0.25em b}\kern-0.8em\TeX}}}

\setcopyright{acmcopyright}
\acmJournal{TOMM}
\acmYear{2021} \acmVolume{1} \acmNumber{1} \acmArticle{1} \acmMonth{1} \acmPrice{15.00}\acmDOI{10.1145/3495211}

\usepackage{wrapfig}
\usepackage{multirow}

\usepackage{booktabs}
\newcommand{\mysubsection}[1]{\vspace{0.3em}\noindent\textbf{#1}}
\usepackage{pifont}
\newcommand{\xmark}{x}
\graphicspath{ {img/} }
\usepackage[capitalize,noabbrev]{cleveref}
\usepackage{subcaption} % for subtable

\begin{document}

\setcopyright{acmcopyright}
\acmJournal{TOMM}
\acmYear{2021} \acmVolume{1} \acmNumber{1} \acmArticle{1} \acmMonth{1} \acmPrice{15.00}\acmDOI{10.1145/3495211}

\title{When Did It Happen? Duration-informed Temporal Localization of Narrated Actions in Vlogs}

\author{Oana Ignat}
\email{oignat@umich.edu}
\orcid{0000-0003-0272-5147}
\author{Santiago Castro}
\email{sacastro@umich.edu}
\author{Yuhang Zhou}
\email{tonyzhou@umich.edu}
\author{Jiajun Bao}
\email{jiajunb@umich.edu}
\author{Dandan Shan}
\email{dandans@umich.edu}
\author{Rada Mihalcea}
\email{mihalcea@umich.edu}
\affiliation{%
  \institution{University of Michigan}
  \streetaddress{500 S State St}
  \city{Ann Arbor}
  \state{Michigan}
  \country{USA}
  \postcode{48109}
}

\renewcommand{\shortauthors}{Ignat, et al.}

\begin{abstract}
 We consider the task of temporal human action localization in lifestyle vlogs.
We introduce a novel dataset consisting of manual annotations of temporal localization for 13,000 narrated actions in 1,200 video clips. We present an extensive analysis of this data, which allows us to better understand how the language and visual modalities interact throughout the videos. We propose a simple yet effective method to localize the narrated actions based on their expected duration. Through several experiments and analyses, we show that our method brings complementary information with respect to previous methods, and leads to improvements over previous work for the task of temporal action localization.
\end{abstract}


%   <concept>
%       <concept_id>10010147.10010178.10010179</concept_id>
%       <concept_desc>Computing methodologies~Natural language processing</concept_desc>
%       <concept_significance>500</concept_significance>
%       </concept>
%   <concept>
%       <concept_id>10010147.10010178.10010224</concept_id>
%       <concept_desc>Computing methodologies~Computer vision</concept_desc>
%       <concept_significance>500</concept_significance>
%       </concept>
%   <concept>
%       <concept_id>10010147.10010178.10010224.10010225.10010228</concept_id>
%       <concept_desc>Computing methodologies~Activity recognition and understanding</concept_desc>
%       <concept_significance>500</concept_significance>
%       </concept>
%  </ccs2012>

% \ccsdesc[500]{Computing methodologies~Natural language processing}
% \ccsdesc[500]{Computing methodologies~Computer vision}
% \ccsdesc[500]{Computing methodologies~Activity recognition and understanding}


\maketitle

\section{Introduction}
Targetting the long-term goal of video understanding, recent years have witnessed significant progress in the task of action localization, starting with the localization of one action at a time in a short clip \cite{Wang2011ActionRB} or in a longer untrimmed video \cite{lin2017single}, all the way to localizing more complex natural language queries in videos \cite{gao2017tall, Hendricks_2017_ICCV, hendricks2018localizing, ge2019mac, ghosh2019excl}, and recently to localizing complex natural language queries extracted directly from transcripts in online videos \cite{tang2019coin, zhukov2019cross, miech2019end}.

Lifestyle vlogs represent a great challenge and opportunity for this task, as they depict everyday actions in a complex setting. Unlike traditional action datasets \cite{%malmaud2015s,
caba2015activitynet, sigurdsson2016hollywood, abu2016youtube, Hendricks_2017_ICCV} or instructional video datasets \cite{zhukov2019cross, tang2019coin, miech2019howto100m}, vlogs contain a wide variety of actions that are more akin to real-life settings, such as ``grab my Kindle,'' ``do some reading,'' or ``chill out.''

Moreover, vlogs typically include transcripts with complex natural language expressions, which allow us to find an alternative to the costly process of manual annotations. Given the prevalence of vlogs in online platforms, automatically extracting  action names from their transcripts can lead to a large-scale inexpensive action dataset.  Previous work \cite{miech2019howto100m} relied on this technique to build very large datasets of video-action mappings. However, previous work also found that the video and transcript are often misaligned  \cite{Ignat2019IdentifyingVA, miech2019end}: in the best case, there is a gap of a few seconds between the time when a person verbally expresses the action and when it is visually illustrated.

\begin{figure*}
    \includegraphics[width=1\textwidth]{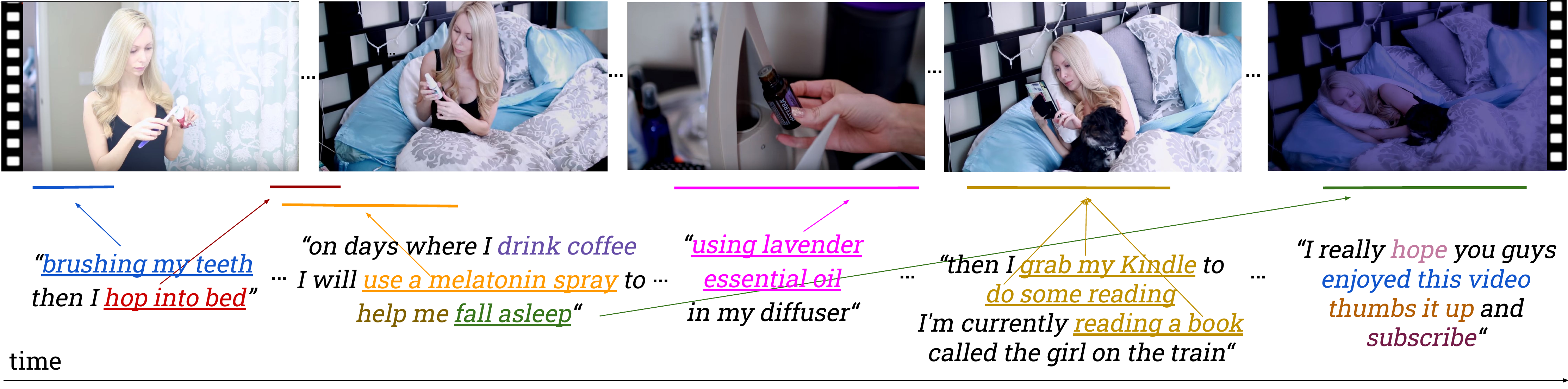}
    \caption{Overview of the dataset \cite{Ignat2019IdentifyingVA}:
    distinguishing between actions that are narrated by the vlogger but not visible in the video and actions that are both narrated and visible in the video (underlined), with a highlight on visible actions that represent the same activity (same color). The arrows represent the temporal alignment between when the visible action is narrated as well as the time it occurs in the video. Best viewed in color.}
    \label{fig:routine_example}
\end{figure*}

This paper addresses the task of temporal action localization in vlogs, and makes three main contributions. 
First, we introduce a dataset of manual annotations of temporal localization of actions that addresses new challenges compared to other action localization datasets. 
Second, we present {\sc 2Seal} -- a simple yet effective method that leverages both language and vision to temporally localize actions, while also accounting for the expected duration of the actions. Through extensive evaluations, we show that our proposed method can be used along with existing models to improve their performance on temporal action localization. Finally, we conduct an analysis of the results, and gain insight into the role played by the different components, which further suggests avenues for future work.

\section{Related Work}

Learning connections between vision and language is crucial to many  applications. These applications include visual question answering \cite{Yang2015StackedAN, anderson2018bottom, Lei2019TVQASG}, visual content retrieval based on textual queries \cite{miech2019howto100m,mithun2018learning, Jiang2007TowardsOB}, image and video captioning \cite{donahue2015long, you2016image, anderson2018bottom}, video summarization with natural language \cite{palaskar2019multimodal, plummer2017enhancing}, action detection \cite{carreira2017quo, lin2017single, girdhar2019video}, action temporal localization in videos \cite{regneri-etal-2013-grounding, gao2017tall, ghosh2019excl, escorcia2019temporal, ge2019mac} and mapping text descriptions to image or video content \cite{%rohrbach2013translating,
regneri-etal-2013-grounding, wang2016structured, lee2018stacked, shi2019dense,krishna2017dense}.

\paragraph{Action Localization Datasets.}

Action detection and localization algorithms evolve with the building of complex datasets. From searching YouTube videos, given a set of predefined actions \cite{Heilbron2015ActivityNetAL, abu2016youtube, carreira2017quo}, or filming in people's homes who act based on a scenario \cite{sigurdsson2016hollywood}, these datasets capture the complexity of daily life activities. However, because of the high annotation cost, these methods are not scalable. Currently, the latest trend in the vision community is to search for pre-defined tasks on WikiHow and collect their corresponding videos from YouTube \cite{miech2019howto100m, tang2019coin, zhukov2019cross}. This process is more efficient and guarantees that more relevant actions are shown in the videos.  
Another technique for collecting human actions is to perform implicit data gathering \cite{fouhey2018lifestyle}: instead of explicitly searching for a pre-defined task, find routine videos that contain a broad range of daily actions.

In our work, we use the data introduced in \cite{Ignat2019IdentifyingVA} which identifies if the actions mentioned in the transcripts are present (visible) in the video. Although we use implicit data gathering as proposed in the past, unlike Fouhey et al. \cite{fouhey2018lifestyle},  who focus on the visual information (hand and object locations), we focus on routine videos that contain rich audio descriptions of the actions being performed, and we use this transcribed audio to extract actions.

\paragraph{Action Localization Methods.}

Methods that reason over text and visual information do this by first extracting the textual embeddings \cite{Pennington2014GloveGV, Levy2014DependencyBasedWE, Devlin2019BERTPO} and visual features \cite{Tran2014C3DGF, carreira2017quo} and then linearly mapping them to the same embedding space \cite{anne2017localizing, Hendricks_2017_ICCV, gao2017tall, ge2019mac}. 
This is usually computed using self and cross attention over the textual and visual features. The visual features can be extracted with a convolutional neural net as in \cite{anne2017localizing, Yuan2018ToFW, Liu2018AttentiveMR} or from object bounding boxes \cite{lee2018stacked}. 
Recent work \cite{tan2019lxmert, sun2019videobert, lu2019vilbert} builds on this approach by combining the attention modules in a large scale Transformer architecture \cite{Vaswani2017AttentionIA}. Their goal is to learn inter-modality and cross-modality relationships that can be used in downstream tasks that require complex reasoning about natural language grounded in visual data \cite{Goyal2016MakingTV, suhr2018corpus, hudson2019gqa}. 

\paragraph{Instructional vs. Routine Videos.}

Action localization methods are moving from using simple pre-defined action labels \cite{carreira2017quo, girdhar2019video} to more complex natural language action descriptions \cite{
anne2017localizing, sigurdsson2018charades, miech2019howto100m}. Our goal is also to localize natural language descriptions of actions in videos. An important difference between our task and previous work is that the natural language descriptions come from the people filming the actions. 

Research work such as \cite{alayrac2016unsupervised, miech2019howto100m} also take advantage directly of the actions extracted from the transcripts, however their videos are instructional videos. Instead of looking at instructional videos, we choose a broader category: routine videos, which can contain instructions, but are more focused on describing the typical day of a person.

Compared to instructional videos, routine videos contain a more diverse set of activities, from waking up in the morning and taking a shower, to working out and making a meal. This diversity of actions in one video translates to many more diverse filming perspectives in the same video, which presents a novel challenge for action localization models. Another difference is that routine videos contain higher-level actions that can be abstract in nature (e.g., ``wind down,'' ``go for a walk'') and thus harder to ground than clear instructions. This is an important difference, as it presents a challenge that is essential for webly supervised systems, which are expected to learn from a diverse mix of both concrete actions and high-level abstract actions. In the realm of web videos, instructional videos account for only a small fraction.

Finally, note that existing action localization methods by and large rely on simplifying assumptions (e.g., instructional videos, always visible actions, non-overlapping actions). In contrast, our paper introduces an evaluation that accounts for the additional challenges encountered in online videos.

\section{Data Collection and Annotation}
\label{sec:dataset}

We collect a dataset of routine and do-it-yourself (DIY) videos from YouTube, consisting of  people performing daily activities, such as making breakfast or cleaning the house. These videos also typically include a detailed verbal description of the actions being depicted. We choose to focus on these lifestyle vlogs because they are very popular, with tens of millions having been uploaded on YouTube; \cref{tab:nbresults_search_queries} shows the approximate number of videos available for several routine queries. Vlogs also capture a wide range of everyday activities; on average, we find thirty different visible human actions in five minutes of video.

By collecting routine videos, instead of searching explicitly for actions, we do {\it implicit} data gathering, a form of data collection introduced by Fouhey et al \cite{fouhey2018lifestyle}. Because everyday actions are common and not unusual, searching for them directly does not return many results. In contrast, by collecting routine videos, we find many everyday activities present in these videos.

\begin{table}
\parbox{.49\linewidth}{
    \centering
    \begin{tabular}{lc}
    \hline
    Query&Results\\
    \midrule
    my morning routine & 28M+ \\
    my after school routine & 13M+ \\
    my workout routine & 23M+ \\
    my cleaning routine & 13M+ \\
    DIY & 78M+ \\ 
    \hline
    \end{tabular}
    \caption{Approximate number of videos found when searching for routine and do-it-yourself queries on YouTube.}
    \label{tab:nbresults_search_queries}
}
\hfill
\parbox{.49\linewidth}{
 \centering
    % \scalebox{0.85}{
        \begin{tabular}{rc}
            \toprule
            Videos & 171 \\
            Video hours & 20 \\
            Transcript words & 302,316 \\
            Clips & 1,246 \\
            % Actions & 12,533 \\
            Actions & 13,380 \\
            Visible actions & 3,131 \\
            % Visible actions & 4,340 \\
            % Non-visible actions & 9,394 \\
            Non-visible actions & 10,249 \\
            \bottomrule
        \end{tabular}
    % }
    \caption{Data statistics}
    \label{tab:data_stats}
}
\end{table}

\begin{figure*}
    \includegraphics[width=0.5\textwidth]{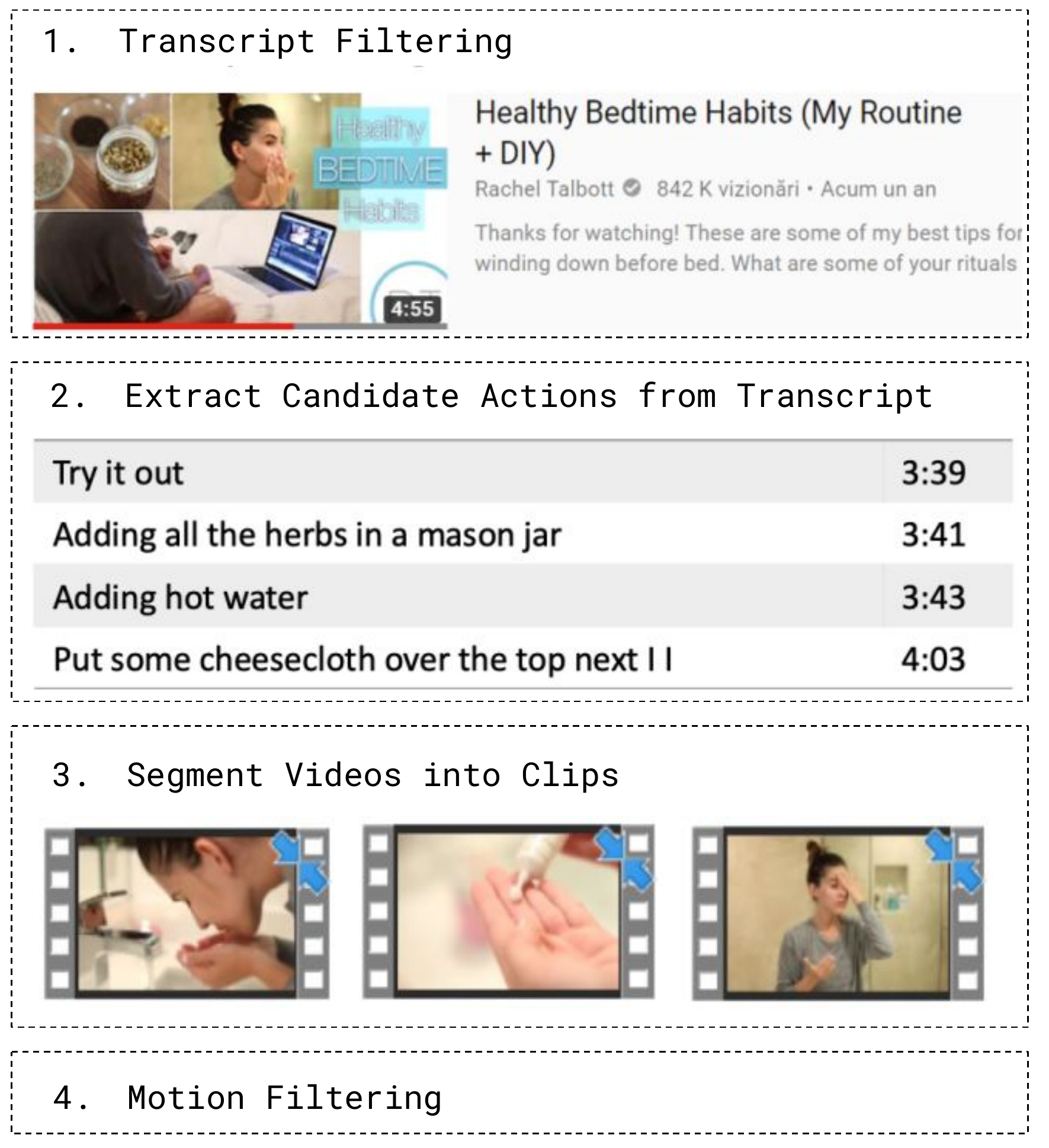}
    \caption{Overview of the data gathering pipeline.}
    \label{fig:dataGathering}
\end{figure*}

\subsection{Data Gathering}
We build a data gathering pipeline (see \cref{fig:dataGathering}) to automatically extract and filter videos and their transcripts from YouTube. The input to the pipeline is manually selected YouTube channels. Ten channels are chosen for their rich routine videos, where the actor(s) describe their actions in great detail. From each channel, we manually select two different playlists, and from each playlist, we randomly download ten videos. The following data processing steps are applied:

\mysubsection{Transcript Filtering.} 
Transcripts are automatically generated by YouTube.
We filter out videos that do not contain any transcripts or that contain transcripts with an average (over the entire video) of less than 0.5 words per second.

These videos do not contain detailed action descriptions so we cannot effectively leverage textual information. 

\mysubsection{Extract Candidate Actions from Transcript.}
Starting with the transcript, we generate a noisy list of potential actions. This is done using the Stanford parser \cite{chen2014fast} to split the transcript into sentences and identify verb phrases, augmented by a set of hand-crafted rules to eliminate some parsing errors. The resulting actions are noisy, containing phrases such as ``found it helpful if you'' and ``created before up the top you.'' 

\mysubsection{Segment Videos into Clips.} The length of our collected videos varies from two minutes to twenty minutes. To ease the annotation process, we split each video into clips (short video sequences of maximum one minute). Clips are split to minimize the chance that the same action is shown across multiple clips. This is done automatically, based on the transcript timestamp of each action.
Because YouTube transcripts have timing information, we are able to line up each action with its corresponding frames in the video. We sometimes notice a gap of several seconds between the time an action occurs in the transcript and the time it is shown in the video. To address this misalignment, we first map the actions to the clips using the time information from the transcript. We then expand the clip by 15 seconds before the first action and 15 seconds after the last action. This increases the chance that all actions will be captured in the clip.

\mysubsection{Motion Filtering.} We remove clips that do not contain significant movement. We sample one out of every one hundred frames of the clip, and compute the 2D correlation coefficient between these sampled frames. If the median of the obtained values is greater than a certain threshold (we choose 0.8), we filter out the clip.

Videos with low movement tend to show people sitting in front of the camera, describing their routine, but not acting out what they are saying. There can be many actions in the transcript, but if they are not depicted in the video, we cannot leverage the video information.

\begin{figure}
    \centering
    \includegraphics[width=0.9\textwidth]{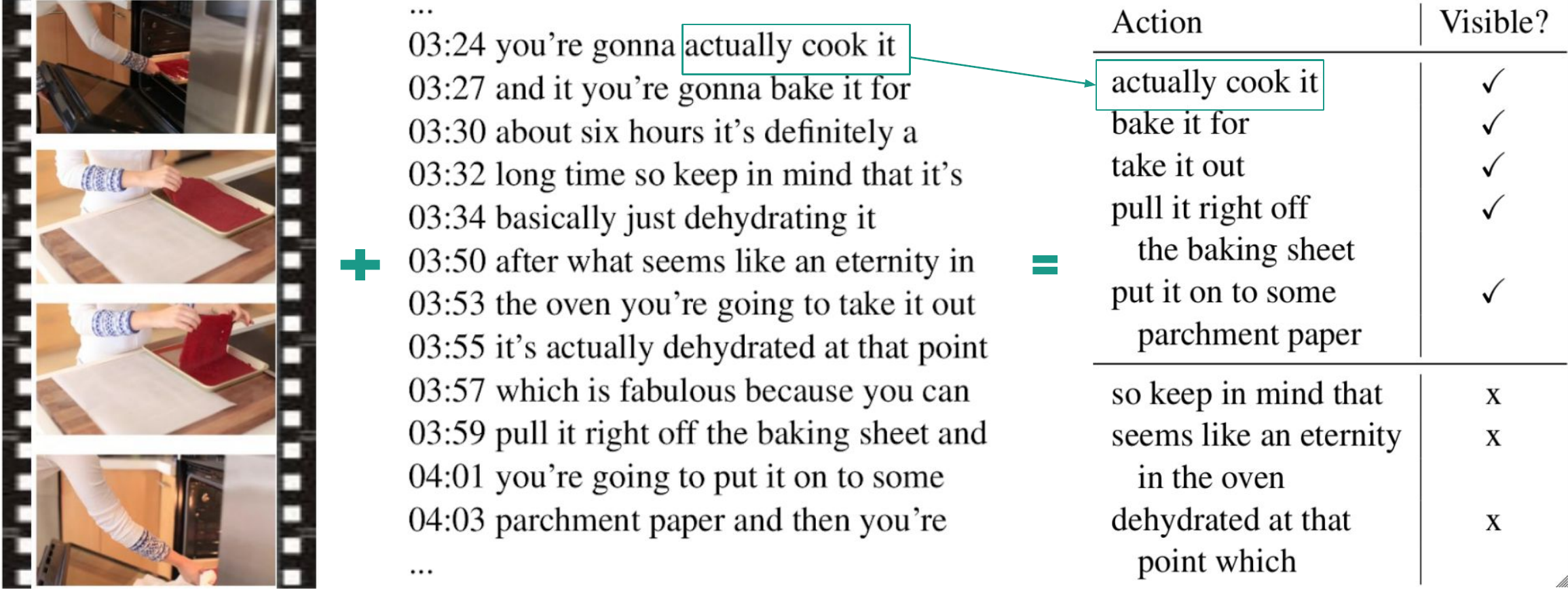}
    \caption{Sample video frames, transcript, and annotations.}
    \label{fig:sample}
\end{figure}

\subsection{Visual Action Annotation}

\begin{figure}
    \centering
    \includegraphics[width=0.7\textwidth]{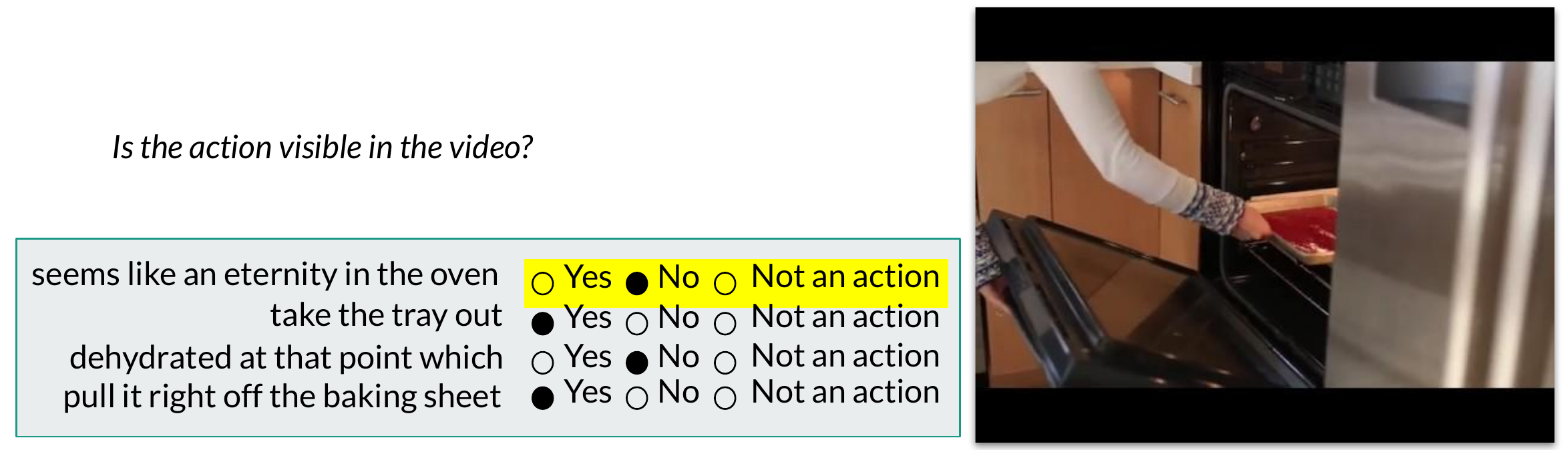}
    \caption{Annotation tool used by Amazon Mechanical Turk workers to annotate if an action is visible or not in the video.}
    \label{fig:annotation_tool_visible}
\end{figure}

We start by identifying which of the actions extracted from the transcripts are visually depicted in the videos. We create an annotation task on Amazon Mechanical Turk (AMT) to identify actions that are visible. 
We give each AMT turker a HIT consisting of five clips with up to seven actions generated from each clip.
The turker is asked to assign a label (\textit{visible} in the video; \textit{not visible} in the video; \textit{not an action}) to each action.
\cref{fig:annotation_tool_visible} shows the AMT interface used.
Because it is difficult to reliably separate \textit{not visible} and \textit{not an action}, we group these labels together. 
Each clip is annotated by three different turkers.  For the final annotation, we use the label assigned by the majority of turkers, i.e., \textit{visible} or \textit{not visible / not an action}.

To help detect spam, we identify and reject the turkers that assign the same label for every action in all five clips that they annotate. Additionally, each HIT contains a ground truth clip that has been pre-labeled by two reliable annotators. Each ground truth clip has more than four actions with labels that were agreed upon by both reliable annotators. 
We compute accuracy between a turker's answers and the ground truth annotations; if this accuracy is less than 20\%, we reject the HIT as spam. 

\begin{figure}
 \centering
    \includegraphics[width=0.7\textwidth]{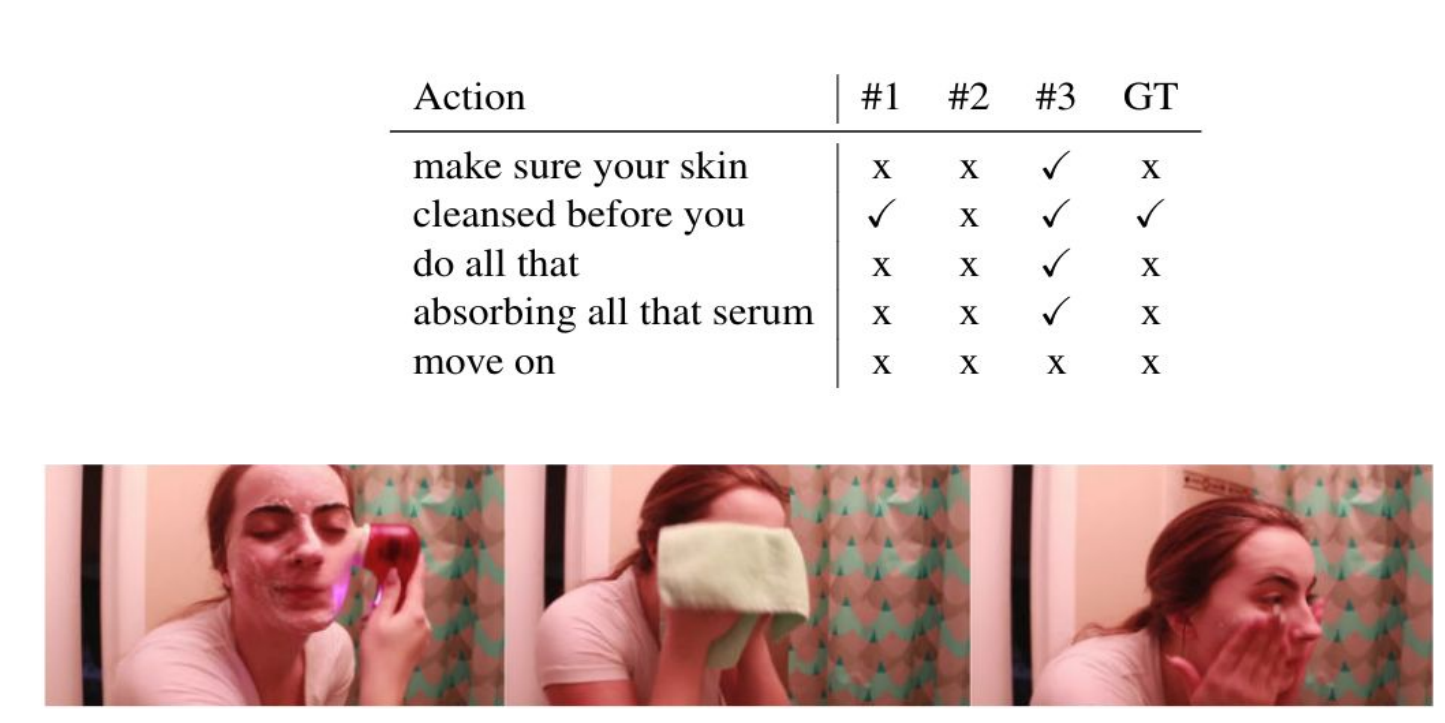}
    \caption{An example of low agreement. The table shows actions and annotations from workers \#1, \#2, and \#3, as well as the ground truth (GT). Labels are: visible - \checkmark, not visible - x. The bottom row shows screenshots from the video. The Fleiss kappa agreement score is -0.2.}
    \label{fig:example_low_agreement}
\end{figure}

After spam removal, we compute the agreement score between the turkers using Fleiss kappa \cite{fleiss1973equivalence}. Over the entire  data set, the Fleiss agreement score is 0.35, indicating fair agreement. On the ground truth data, the Fleiss kappa score is 0.46, indicating moderate agreement. This fair to moderate agreement indicates that the task is difficult, and there are cases where the visibility of the  actions is hard to label. To illustrate, Figure \ref{fig:example_low_agreement} shows examples where the annotators had low agreement.
Table \ref{tab:data_stats} shows statistics for our final dataset of videos labeled with actions, and \cref{fig:sample} shows a sample video and transcript, with annotations.

% Previous work\cite{Ignat2019IdentifyingVA} introduced the task of detecting whether an action mentioned in the video is also visible in the video. To annotate the actions as visible or not visible, Ignat et al. \cite{Ignat2019IdentifyingVA} first split the videos into smaller clips of approximately one minute, based on the timing information present in the transcript. The visibility of the actions was annotated with the help of Amazon Mechanical Turk (AMT): instead of annotating a whole video, each worker received a short clip with a list of actions that had to be binary labeled as \textit{visible} or \textit{not visible} in the clip. Each clip contains on average fifteen actions, out of which five are \textit{visible}. 

% Overall statistics of the dataset are shown in  \cref{tab:data_stats}. Further details about the dataset can be found in \cite{Ignat2019IdentifyingVA}.

\begin{figure}
    \centering
    \includegraphics[width=0.9\textwidth]{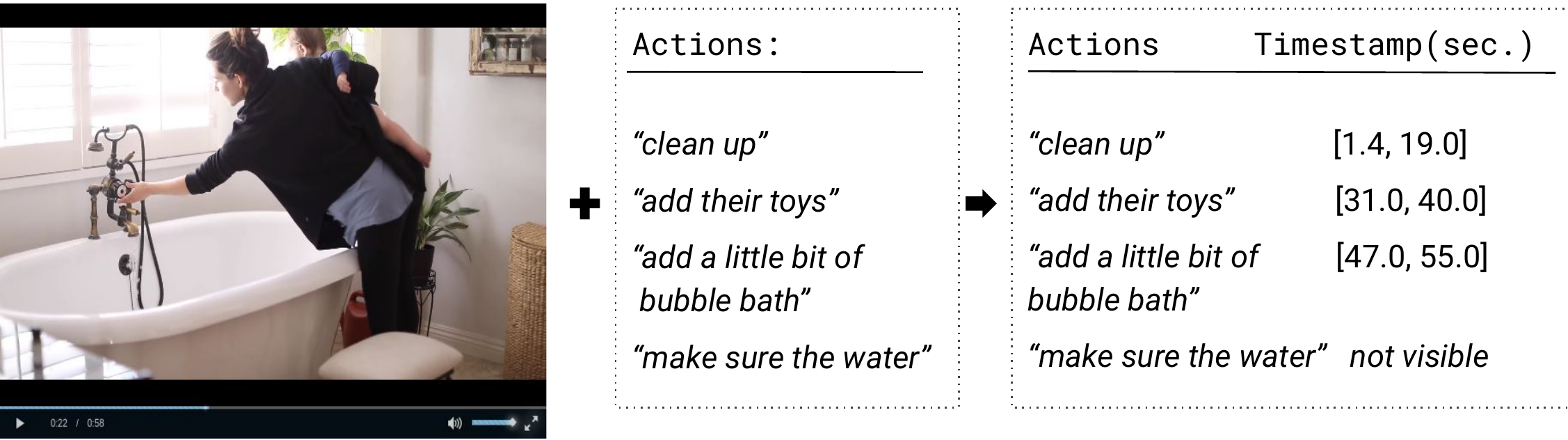}
    \caption{Action temporal localization annotation. Each action is localized in the video according to its start and end time offsets. The action is localized according to its visibility in the video, and if it cannot be seen, it is marked as \textit{not visible}.}
    \label{fig:annotation_example}
\end{figure}

Note that the goal of our dataset is to capture naturally-occurring, routine actions. Because the same action can be identified in different ways (e.g., ``pop into the freezer'', ``stick into the freezer"), our dataset has a complex and diverse set of action labels. These labels demonstrate the language used by humans in everyday scenarios; because of that, we choose not to group our labels into a pre-defined set of actions. Table \ref{tab:comparison_statistics} shows the number of unique verbs, which can be considered a lower bound for the number of unique actions in our dataset. On average, a single verb is used in seven action labels, demonstrating the richness of our dataset.

The action labels extracted from the transcript are highly dependent on the performance of the constituency parser. This can introduce noise or ill-defined action labels. Some actions contain extra words (e.g., ``brush my teeth of course''), or lack words (e.g., ``let me just''). Some of this noise is handled during the annotation process; for example, most actions that lack words are labeled as ``not visible'' or ``not an action'' because they are hard to interpret.

\begin{table*}\setlength{\tabcolsep}{3pt}
    \centering
    \begin{tabular}{l c c c c c c c}
    \toprule
        Dataset  & \#Actions & \#Verbs& \#Actors & Implicit & Label types\\
        \midrule
       Ours & 4340 & 580 & 10  & \checkmark & \checkmark  \\
        \midrule
        VLOG  \small{\cite{fouhey2018lifestyle}} &- & - & 10.7k  & \checkmark & \checkmark \\
        Kinetics \small{\cite{kay2017kinetics}}&600& 270 & - & \xmark & \xmark \\
        ActivityNet \small{\cite{caba2015activitynet}} &203 & - & -  & \xmark & \xmark \\
        MIT \small{\cite{monfort2019moments}}&339& 339 & - & \xmark & \xmark \\
        AVA \small{\cite{gu2018ava}}&80&80 & 192  & \checkmark & \xmark \\
        Charades \small{\cite{sigurdsson2016hollywood}}&157 & 30 &267  & \xmark & \xmark \\
        MPII Cooking \small{\cite{rohrbach2012database}}&78 &78 & 12 & \checkmark & \xmark \\
        \bottomrule
    \end{tabular}
    \caption{Comparison between our dataset and other video human action recognition datasets. \# Actions show either the number of action classes in that dataset (for the other datasets), or the number of unique visible actions in that dataset (ours); \# Verbs shows the number of unique verbs in the actions; Implicit is the type of data gathering method (versus explicit); Label types are either post-defined (first gathering data and then annotating actions): \checkmark, or pre-defined (annotating actions before gathering data): \xmark.}
    \label{tab:comparison_statistics}
    
\end{table*}

\subsection{Temporal Action Annotation}

Each video is associated with a set of human actions, in the form of verb phrases extracted from the automatically generated video transcripts. The actions are labeled into two categories: \textit{visible} or \textit{not visible}, depending on whether the actions are explicitly represented in the video. For example, in the video sequence shown in \cref{fig:routine_example}, the action ``drink coffee'' is \textit{not visible} in the video; it is only mentioned as a reason for performing the \textit{visible action} of ``use a melatonin spray.'' Other \textit{not visible} actions from \cref{fig:routine_example} are: ``help,'' ``hope,'' ``enjoyed this video,'' ``thumbs it up'' and ``subscribe,'' which relate to video feedback but are not visually shown.

Two of the authors of this paper annotated the start and end time of all the {\it visible} actions in the dataset, as illustrated in \cref{fig:annotation_example}.  Each action is localized according to its start and end time offsets. The timestamp is marked according to when the action is visible, which does not necessarily correspond to when it is talked about.  If the annotators were not able to localize the action in the clips, they marked it as \textit{not visible}, which corresponds to a correction of the original dataset \cite{Ignat2019IdentifyingVA}. They performed the annotations using a simple  annotation tool that we built for this purpose, which is publicly available at \url{https://github.com/OanaIgnat/video_annotations}.

We measure the inter-annotator agreement by computing the Krippendorff's Alpha score \cite{krippendorff1970estimating} using the interval difference function for each video. We obtain scores between 0.78 and 0.90, which indicate a high agreement.

For our experiments, we split the data by vlog channel. Out of ten channels, six channels are used for training, two channels for validation, and two for testing. Statistics for this experimental split are shown in \cref{tab:train-test-eval-split}.

\begin{table*}
\centering
%  \scalebox{0.9}{
\setlength{\tabcolsep}{0.8em} % for the horizontal padding
{\renewcommand{\arraystretch}{1.1}% for the vertical padding

\begin{tabular}{lcccc}
    %   &   & &  \\%   & $\sharp$mini- & $\sharp$(action, \\
    & \#actions & Vis.\ (\%) & \#videos & \#clips \\ %& clips & miniclip)\\
    \toprule
    %   Train & 1,761 & 108 & 549 \\
    Train & 4,939 & 35.1 & 110 & 680\\
    %   & 21,876 & 45,014\\
    %   Val & 466 & 24 & 136 \\
    Val & 1,264 & 35.9 & 26 & 187 \\
    %   & 4,738 & 10,106\\
    %   Test & 904 & 35 & 228 \\
    Test & 3,456 & 25.7 & 35 & 275 \\
    %   & 11,654 & 46,671 \\
    %   \bottomrule
    \end{tabular}
 }
\caption{Statistics for the experimental data split. ``Vis.'' is the percentage of visible actions among the narrated actions.}
\label{tab:train-test-eval-split}
\end{table*}

\subsection{Data Analysis}%
\label{sec:dataanalysis}

We perform two types of analyses to gain a better understanding of our dataset. 

\mysubsection{Action Duration.}

First, we measure the distribution of action durations in our dataset. As shown later, this information is important, as the action durations can have an impact on the performance of different models. \Cref{tab:data_duration_stats1} shows the action duration distribution in the dataset. A summary of {\it long} actions found in other datasets is shown in \cref{tab:data_duration_stats3} (we define an action as \textit{long} if it exceeds fifteen seconds). \Cref{tab:data_duration_stats2} shows examples of \textit{long} actions, grouped by verb and sorted by frequency.

\begin{table}
    \begin{subtable}[t]{0.45\linewidth}
        \centering
        % \scalebox{0.8}{
            \begin{tabular}{lr}
            % \hline
            Duration (s) & \#actions \\
            \midrule
                 0-5 & 1,136 \\
                 5-15 & 1,200 \\
                 15-25 & 475 \\
                 25-35 & 157 \\
                 35-45 & 72 \\
                 45-60 & 99 \\
            % \hline
            \end{tabular}
        % }
        \caption{}
        \label{tab:data_duration_stats1}
        \end{subtable}%
        \begin{subtable}[t]{0.55\linewidth}
        \centering
        % \scalebox{0.8}{
            \begin{tabular}{lr}
                Long actions & \#actions \\
                \midrule
                use (a whisk) & 87 \\
                make (oatmeal) & 81\\
                clean (my skin) & 60\\
                \midrule
                \midrule
                Short actions & \#actions \\
                \midrule
                add (spice) & 362\\ 
                use (the clamps) & 228\\
                put (a lid on top) & 179\\
            \end{tabular}
        % }
        \caption{}
        \label{tab:data_duration_stats2}
        \end{subtable}
        \begin{subtable}[t]{\textwidth}
        \centering
        % \scalebox{0.85}{
            \begin{tabular}{lr}
                Dataset & Long actions (\%) \\
                \midrule
                Charades-STA \cite{gao2017tall} & 4.2 \\
                CrossTask \cite{zhukov2019cross} & 16.4 \\
                COIN \cite{tang2019coin} & 31.6 \\
                % HowTo100M \cite{miech2019howto100m} & 1.0 \\
                \midrule
                Ours & 25.5 \\
            \end{tabular}
        % }
        \caption{} %Number of \textit{long} actions relative to total number of actions in related datasets. MIL-NCE data is based on their provided transcript data.}
        \label{tab:data_duration_stats3}
    \end{subtable}%
    \label{tab:data}
    \caption{Action duration analysis: (a) Distribution in our dataset; (b) Example of long and short actions, each with a sample object, grouped by verbs and sorted by verb frequency; (c) Percentage of long ($>$15s) actions in other datasets.}
\end{table}

\mysubsection{Temporal Relations between Actions.}

Second, we analyze the temporal relations between actions mentioned in the transcripts. These actions can be challenging to model as they capture the complexities of real life.
While there are several actions that follow each other (as more naturally expected), there are also actions that overlap, are included in one another, or even happen at the same time.
From a total of 2,070 number of overlapping actions, 1,573 are included in each other and 269 occur exactly at the same time.
\Cref{tab:data_stats_overlapping} shows examples of such actions.
While several action localization datasets have been proposed in the past \cite{tang2019coin}, to the best of our knowledge, this dataset is the only action localization dataset that contains {\it overlapping} actions, making it challenging and novel.
For the purpose of this work, we localize each action independent of other actions, but future work may leverage the relations that exist between actions.

% \begin{table}
%     \centering
%     % \resizebox{\columnwidth}{!}{%
%         \begin{tabular}{c}
%             \textit{``toss everything together''} $\cap$ \textit{``chop it up''} \\
%             \textit{``add fresh herbs''} $\cap$ \textit{``add chickpeas to a bowl''} \\
%             \textit{``get them dress for bedtime''} $\cap$ \textit{``wind down with both of them''} \\
%             \midrule
%             \textit{``use a plastic scraper''} $\subseteq$ \textit{``wipe thoroughly''} \\ 
%             \textit{``throw the cushions around''} $\subseteq$ \textit{``fix my cushions up''} \\
%             \textit{``take that with a glass of water''} $\subseteq$ \textit{``take my vitamin d''} \\
%             \midrule
%             \textit{``make your bucket list''} $\equiv$ \textit{``write those things down''} \\
%             \textit{``grab my Kindle''} $\equiv$ \textit{``do some reading''} \\
%             \textit{``use one tablespoon of cashew nut butter''} $\equiv$ \textit{``add good protein''} \\
%         \end{tabular}%
%     % }
%     \caption{Examples of different types of action temporal relations: actions that overlap ($\cap$), actions that are included in each other ($\subseteq$), actions that occur exactly at the same time ($\equiv$).}
%     \label{tab:data_stats_overlapping}
% \end{table}

\begin{table}
\resizebox{0.9\columnwidth}{!}{%
{\renewcommand{\arraystretch}{1.5}% for the vertical padding
    \centering
    \begin{footnotesize}
    \begin{tabular}{ c | c }
         Actions that follow each other & Actions that overlap \\
        \hline
        \textit{``make super quick chicken tacos''} ; \textit{``do the dishes"} & \textit{``toss everything together''} $\cap$ \textit{``chop it up''} \\
        \textit{``put them in a bowl''} ; \textit{``cover in water''} & \textit{``add fresh herbs''} $\cap$ \textit{``add chickpeas to a bowl''} \\ 
        \textit{``give a little mix''} ; \textit{``add half cup of berries''} &  \textit{``scoop out of the processor''} $\cap$ \textit{``scoop it into a bowl''} \\
        \textit{``get a little water on your skin''} ; \textit{``rinse it off''} &  \textit{``combine our dry ingredients''} $\cap$ \textit{``give it a mix''}\\
        ... & ... \\
        \midrule
        Actions that are included in each other & Actions that occur exactly at the same time \\
        \hline
        \textit{``use a plastic scraper''} $\subseteq$ \textit{``wipe thoroughly''} &  \textit{``write out''} $\equiv$ \textit{``make your bucket list''} \\ 
        \textit{``throw the cushions around''} $\subseteq$ \textit{``fix my cushions up''} &  \textit{``go to bed''} $\equiv$ \textit{``head to bed''}\\
        \textit{``do this scrub vigorously''} $\subseteq$ \textit{``clean some ovens''} &  \textit{``add good protein''} $\equiv$ \textit{``use one tablespoon of cashew nut butter''}\\
        \textit{``do some yoga''} $\subseteq$ \textit{``wind down''} & \textit{``grab my Kindle''} $\equiv$ \textit{``do some reading''}\\
        ... & ... \\
        
    \end{tabular}
    \end{footnotesize}
    }
    }
    \caption{Examples of different types of action temporal relations: actions that overlap ($\cap$), actions that are included in each other ($\subseteq$), actions that occur exactly at the same time ($\equiv$). From a total of 2,070 number of overlapping actions, 1,573 are included in each other and 269 occur exactly at the same time.}
    \label{tab:data_stats_overlapping}
\end{table}

\section{Two-Stage Action Localization}

For a given action mentioned in a video transcript, our goal is to: (1) decide if it is visible within the video clip; and (2) if it is visible, identify its temporal location (i.e., the time interval start and end times).

To achieve this goal, we propose a two-stage method which we call {\sc 2Seal} (2-StagE Action Localization). 

\begin{figure*}
    \centering
    \includegraphics[width=1\textwidth]{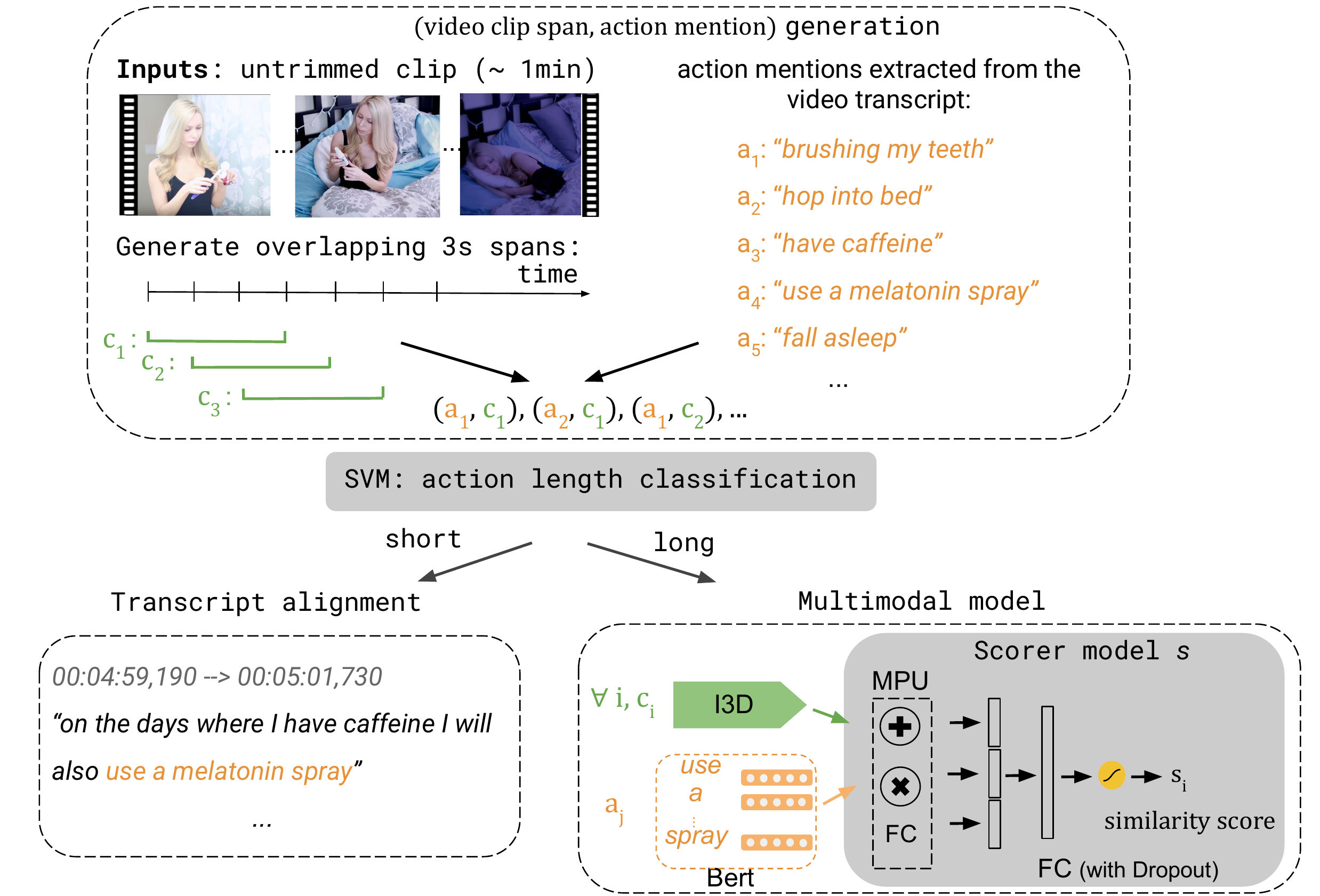}
    \caption{2SEAL method architecture. Note the depicted MPU-based multimodal model can be replaced with any multimodal model. The MPU model is composed of vector element-wise addition ('+'), vector element-wise multiplication ('x') and vector concatenation followed by a Fully Connected ('FC') layer to combine the information from both textual and visual modalities.}
    \label{fig:method_architecture}
\end{figure*}

\Cref{fig:method_architecture} shows the overall architecture of {\sc 2Seal}. Following our analysis of the variation in action duration (see \cref{sec:dataanalysis}), and empirical observations made on the development dataset, we hypothesize that shorter actions can be localized mainly based on the temporal information inferred from the transcript (i.e., \emph{when} an action was narrated within the transcript), whereas longer actions are often temporally shifted with respect to their mention in the transcript and thus can benefit from a multimodal model. We thus devise an architecture that first aims to predict whether the action is short or long, and correspondingly activates a transcript alignment (for short actions) or a multimodal model (for long actions). We describe below each of these main components.

\mysubsection{Action Duration Classification} We use the annotated temporal locations in the videos to determine the expected duration of each action, and build a binary classifier to discriminate between short ($\leq$15s) and long ($>$15s) actions. We choose this threshold based on the validation data.
The classifier uses as input an action text embedding obtained from a text encoder, as described in \cref{sec:experiments}.

\begin{figure*}
    \centering
    \includegraphics[width=0.7\textwidth]{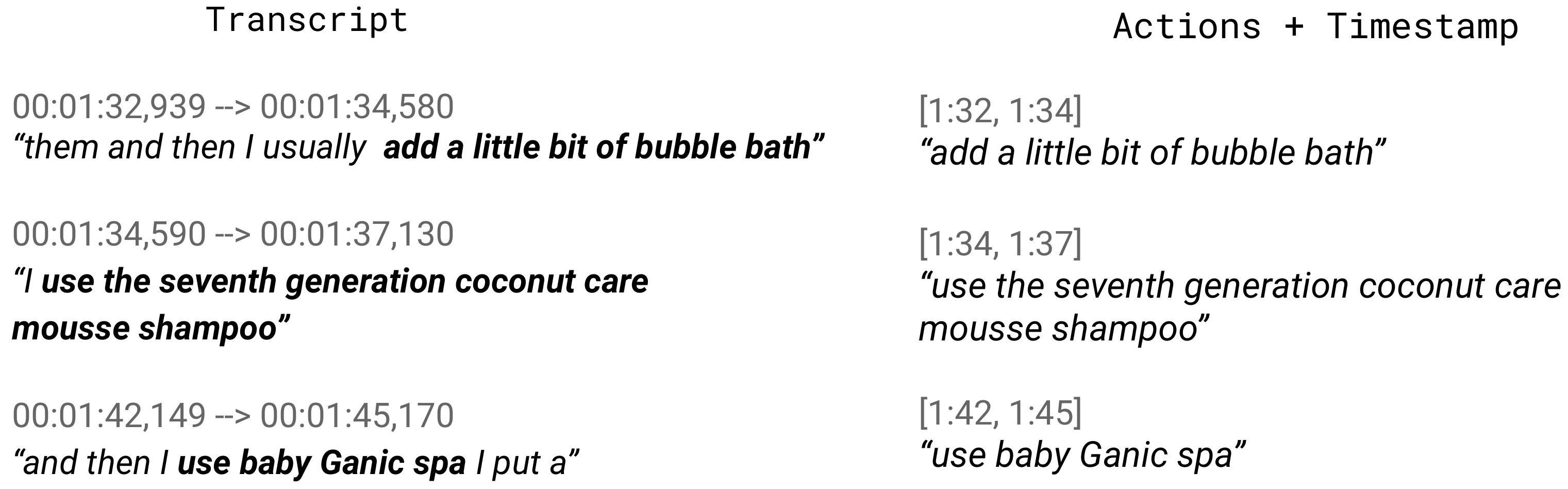}
    \caption{Example of applying the Transcript Alignment method. The transcript contains time intervals for utterances. Each action contained in an utterance is assigned the corresponding time interval.}
    \label{fig:transcript_alignment}
\end{figure*}

\mysubsection{Transcript Alignment.} Each video contains a transcript automatically generated by the YouTube API. The transcript contains time information for every utterance. Given an action mention extracted from an utterance, the Transcript Alignment method assumes the action is visible, and predicts its temporal location to be the time interval associated with the corresponding utterance, as illustrated in \cref{fig:transcript_alignment}. The transcript alignment is also illustrated in \cref{fig:method_architecture}.

\mysubsection{Multimodal Model.} We split the video clips into fixed-duration spans and convert the action temporal localization task into binary classification tasks based on the output from a scorer model \textit{s}.
We aim to predict if the visual information from a video clip span corresponds to the linguistic representation of an action.
For a given action mention within the transcript and a fixed-duration video clip span, we compute a similarity score to decide if they correspond to each other.
The action mention is represented using a text encoder and the features for the video clip span are obtained from a video encoder (see \cref{sec:experiments}).

The process of pairing action mentions to video clip spans is shown in \cref{fig:method_architecture}; \textit{s} can be represented by any multimodal model, and we describe several models in \cref{subsec:main_results}. 
At test time, given a video clip and its corresponding transcript, we input all the pairs of action mentions and fixed-duration video clip spans. We merge all the spans that surpass a certain threshold and are separated by less than three seconds into \emph{proposals}. Each proposal is assigned the maximum similarity score of its spans. We then perform non-maximum suppression to select the best proposal as the predicted action location interval.
At training time, we focus only on the binary task and train \textit{s} with the standard cross-entropy loss. Given that an action mention has many more negative ({\it not visible}) fixed-duration video clip spans in a given video clip, we balance the classes out via downsampling by taking negative random samples from the same video clip.
The question of how different negative sampling strategies affect the scorer model performance is left for future work.

\section{Experiments}%
\label{sec:experiments}

To evaluate our duration-informed action localization method, we run several comparative experiments on the dataset described in \cref{sec:dataset}. We compare our method with several strong baselines, and also perform feature ablation and a breakdown of results by action duration.

In all our experiments, we use a video encoder consisting of the last layer (\texttt{mixed\_5c}) from a Kinetics~\cite{carreira2017quo} pre-trained I3D model. The video clips are divided into overlapping three-second spans with a stride of 1s.
We freeze both the text and the video encoders and take their outputs as features.
For the Action Duration Classification, we use an SVM classifier with C=1.0 and an RBF kernel, and  weight the samples inversely proportional to their class frequency.
We train the models using an Adam optimizer with early stopping (tolerance 15 epochs), with a learning rate of 0.001 and a batch size of 64.

\subsection{Action Duration Classification.}

We train the action duration classifier described in the previous section using only the visible actions. The results are reported in \cref{tab:results_action_duration}.  For comparison, we also show the performance of a majority classifier, which labels every action as ``short" by default. As shown in the table, despite the simplicity of the classifier, the action duration classifier obtains good improvement over the majority baseline.

\begin{table}
\centering
    % \resizebox{\columnwidth}{!}{%
        \begin{tabular}{lrrrr}
            \multicolumn{1}{c}{Method} & \multicolumn{1}{c}{A} & \multicolumn{1}{c}{P} & \multicolumn{1}{c}{R} & \multicolumn{1}{c}{F1} \\
            \midrule
            Majority & 74.4 & 74.4 & 100.0 & 85.3 \\
            Action Duration Clf.\  & 80.6 & 81.8 & 97.6 & 89.0 \\
            %   + COIN augmentation & 79.5 & \textbf{83.5} & 92.8 & 87.9 \\
            %   + domain adaptation + COIN augmentation & 78.4 & 83.3 & 91.5 & 87.2 \\
        \end{tabular}%
    % }
    \caption{Action duration classification results on the validation set. The classification is binary, where the positives are the short actions ($\leq$15s) and the negatives the long ones ($>$15s). The columns are in order: accuracy (A), precision (P), recall (R) and F1 score (F1).}
    \label{tab:results_action_duration}
\end{table}

\subsection{Temporal Action Localization}%
\label{subsec:main_results}
Our {\sc 2Seal} method includes a scorer that measures the similarity between a video clip and an action mention (see \cref{fig:method_architecture}). To implement this scorer, we experiment with three methods proposed in previous work: multimodal processing unit, multiple instance learning noise contrastive estimation, and stacked cross attention.

\mysubsection{Multimodal Processing Unit (MPU).}
We use the MPU model \cite{gao2017tall} to compute the similarity score between the language representation of a narrated action and a video clip span.
For the text features, we fine-tune a pre-trained BERT-base-uncased \cite{Devlin2019BERTPO} for domain adaption by on 884 vlog transcripts with 80,749 sentences. We take embeddings from this model for the action mentions in the transcripts by average pooling (the final embedding size is 768). In \cref{subsec:features} we experiment with variations of this text encoder.
The text and visual features for each pair are linearly mapped to the same embedding space.
Next, the MPU model is applied to compute the interaction between the two vectors of the same duration. The MPU model is composed of vector element-wise addition ('+'), vector element-wise multiplication ('x') and vector concatenation followed by a Fully Connected ('FC') layer to combine the information from both textual and visual modalities. The outputs from all three operations are concatenated to construct a multi-modal representation.
This process is also illustrated in the overall architecture in \cref{fig:method_architecture}.
The resulting representation is given as input to a linear layer and finally to a sigmoid function to obtain a similarity score.

\mysubsection{Multiple Instance Learning Noise Contrastive Estimation (MIL-NCE).} We use the MIL-NCE model from \cite{miech2019end} which was trained on HowTo100M \cite{miech2019howto100m}. The similarity score is computed as a dot product between the text and video encoder outputs. The text encoder takes embeddings from a GoogleNews-pretrained skipgram word2vec~\cite{Mikolov2013EfficientEO} implementation and further processes and pools the embeddings to obtain a fixed-size representation. We use the MIL-NCE I3D\footnote{\url{https://tfhub.dev/deepmind/mil-nce/i3d/1}} visual features, and not the S3D features, for consistency reasons and to ensure a fair comparison between the multimodal models.
We empirically find it beneficial to threshold the similarities at mid-range value after experimenting with linear regression models on the validation data.
Note we do not fine-tune this model but freeze it. Future work can explore how the method benefits from fine-tuning.

\mysubsection{Stacked Cross Attention (SCA).}
We also experiment with the SCA method \cite{lee2018stacked}, and adapt its \textit{Text-Image} formulation. It first attends to image frames with respect to each word, and then compares each word to its corresponding attended frame vector to determine the importance of each word. The relevance $R$ between the $i$-th word and the image is defined as the cosine similarity between the $i$-th word vector $v_i$ and its attended frame vector $a_i^t$.
The final similarity score between image $I$ and sentence $T$ is summarized by average pooling: $S_{A V G}^\prime(I, T) = \frac{1}{n} \sum_{j=1}^{n} R^\prime(e_{j}, a_{j}^{v})$.
The textual features are represented using a Gated Recurrent Unit (GRU) \cite{Graves2008SupervisedSL} as in \cite{lee2018stacked}. 
We use the mid-range threshold for the similarity score.

\mysubsection{2D Temporal Adjacent Networks (2D-TAN).}
We find the 2D-TAN model \cite{2DTAN_2020_AAAI} suitable for our task as it is built to localize multiple natural language queries in a video.

The video clips are represented using C3D \cite{tran2015learning} features and the action queries using GloVe \cite{Pennington2014GloveGV} embeddings, as described in the 2D-TAN paper \cite{2DTAN_2020_AAAI}. We take as final proposal the action localization proposal with the highest score. 

We test the pre-trained model and also fine-tune it on our training and validation data.
We run two model configurations, which were trained on TACoS \cite{regneri-etal-2013-grounding}, namely ``Pool'' and ``Conv'' in our test set. ``Pool'' and ``Conv'' represent max-pooling and stacked convolution respectively, which indicates two different ways for moment feature extraction in the 2D-TAN model. We report the results of fine-tuned ``Conv'' 2D-TAN model, which is the best performing 2D-TAN model configuration on our test dataset. 
\subsection{Results}

We evaluate the predictions made by the action localization methods using two evaluation metrics. First, we compute the Visibility Accuracy (VA) to decide if the method can distinguish between visible and not visible actions. Second, only for the visible actions, we compute the recall at different Intersection over Union (IoU) thresholds: 0.1, 0.3, 0.5 and 0.7. A higher threshold means a stronger constraint on how exact the match between the predicted and the ground truth location needs to be. If the predicted interval has an IoU score with the ground truth greater than the threshold, we consider the prediction as being correct. We also compute the average recall over all IoU values, as the mIoU. Note that if a method predicted that a visible action is non-visible, then the recall score is penalized.

\Cref{tab:results} presents the temporal action localization results on our data. The Transcript Alignment method performs better than the MPU, MIL-NCE, SCA and 2D-TAN methods if we do not previously apply our proposed {\sc 2Seal} method before. However, when using our {\sc 2Seal} method that combines both the Transcript Alignment and a method to score long actions (either MPU, MIL-NCE, SCA, or 2D-TAN), the performance improves significantly, with the system integrating the MPU model leading to the best results. We suspect MIL-NCE may perform better if fine-tuned, however our intention is not to compare MPU and MIL-NCE but to show how our method can improve over other existing methods. The results confirm our initial hypothesis that actions of different duration benefit from different methods: the transcript alignment excels at \textit{short} actions, while the multimodal model performs better for \textit{long} actions.

\begin{table*}
\centering
% \scalebox{0.95}{
    \begin{tabular}{l|r|rrrrr}
        \multicolumn{1}{c}{} & \multicolumn{1}{c}{} & \multicolumn{5}{c}{Recall} \\
        % \cline{3-7}
        \multicolumn{1}{c}{Method} & \multicolumn{1}{c}{VA} & \multicolumn{1}{c}{IoU=0.1} & \multicolumn{1}{c}{IoU=0.3} & \multicolumn{1}{c}{IoU=0.5} & \multicolumn{1}{c}{IoU=0.7} & \multicolumn{1}{c}{mIoU} \\
        \midrule
        All visible & 25.7 & 67.4 & 23.6 & 8.3 & 4.1 & 21.6 \\
        All non-visible & 74.3 & 0.0 & 0.0 & 0.0 & 0.0 & 0.0 \\
        \midrule
        Transcript Alignment (ours) & 25.7 & 73.3 & 47.3 & 22.2 & 7.2 & 30.8 \\
        \midrule
        MPU & 75.5 & 57.9 & 27.0 & 12.4 & 6.2 & 21.4 \\
        2SEAL (ours) + MPU & \textbf{79.0} & \textbf{74.6} &  \textbf{48.7} &  \textbf{22.8} & \textbf{8.6} & \textbf{31.9} \\
        \midrule 
        MIL-NCE & 26.1 & 62.9 & 22.2 & 8.0 & 4.2 & 20.5 \\
        2SEAL (ours) + MIL-NCE & 34.4& 74.4 & 47.8 & 21.7 & 7.9 & 31.4  \\
        \midrule 
        SCA & 24.2 & 49.9 & 17.0 & 6.0 & 3.4 & 15.9 \\
        2SEAL (ours) + SCA & 26.1 & 72.2 & 46.7 & 21.4 & 7.6 & 30.5  \\
        \midrule
        2D-TAN & 25.7 & 49.4 & 23.1 & 10.9 & 3.7 & 17.6 \\
        2SEAL (ours) + 2D-TAN & 25.7 & 73.4 & 47.0 & 21.6 & 7.7 & 30.8  \\
        \midrule
        Human &  85.9 & 83.5 & 71.8 & 52.0 & 35.0 & 50.3 \\
    \end{tabular}
% }
\caption{Results on the test set. ``VA'' stands for Visibility Accuracy.}
\label{tab:results}
\end{table*}

\section{Analyses and Discussion}

To gain insights into the performance of our proposed model in relation to action duration, and to understand the role played by different features, we perform several analyses. 
\subsection{Action Duration Impact}%
\label{subsec:action_duration_impact}

If the action is brief, the IoU metric will be  influenced by a few seconds compared to when the action is longer in duration.
This metric penalizes more the mislocalization of short actions, as compared to the longer ones. This analysis is often done for the task of object detection, where the IoU scores are grouped by bounding box size \cite{Redmon2015YouOL}.
To verify our initial hypothesis that actions of different duration benefit from different localization methods, we break down the results of the MPU (the best scorer from among MPU, MIL-NCE, SCA and 2D-TAN without applying the 2SEAL method) by action duration in \cref{tab:results_breakdown_action_duration}. As shown in the table, the performance of the model is connected to the duration of the actions.  For \textit{long} actions, the multimodal method obtains better results compared to the transcript alignment method, while the opposite is true for \textit{short} actions. 

\begin{table}
\centering
% \resizebox{\columnwidth}{!}{%
\setlength{\tabcolsep}{0.4em} % for the horizontal padding
{\renewcommand{\arraystretch}{1.1}% for the vertical padding
    \begin{tabular}{r|r r|r r|r r}
        \multicolumn{1}{l}{} & \multicolumn{2}{c}{0--15s} & \multicolumn{2}{c}{16--35s} & \multicolumn{2}{c}{36--60s} \\
        \cline{2-3} \cline{4-7}
        \multicolumn{1}{c|}{Recall} & MPU & \multicolumn{1}{c|}{Align} & \multicolumn{1}{c}{MPU} & \multicolumn{1}{c|}{Align} & \multicolumn{1}{c}{MPU} & \multicolumn{1}{c}{Align} \\
        \midrule
        IoU=0.1 & 49.5 & \textbf{71.6} & \textbf{90.7} & 76.6 & \textbf{95.2} & 83.3 \\
        IoU=0.3 & 5.4 & \textbf{49.0} & \textbf{73.4} & 51.4 & \textbf{81.0} & 0.0 \\
        IoU=0.5 & 2.0 & \textbf{25.0} & \textbf{22.0} & 17.8 & \textbf{78.6} & 0.0 \\
        IoU=0.7 & 0.8 & \textbf{9.4} & \textbf{5.6} & 1.9 & \textbf{66.7} & 0.0 \\
        \midrule
        mIoU & 12.0 & \textbf{32.0} &\textbf{38.9} & 29.9 &\textbf{71.7} & 16.5 \\
    \end{tabular}
    }%
    % }
    \caption{Breakdown by action duration (time span) on the validation set. The MPU model performance increases with the increase of action time span, while transcript alignment (Align) performance decreases.}
    \label{tab:results_breakdown_action_duration}
\end{table}

% Leave this figure here so it doesn't go to the references section.
\begin{figure}
\centering
    \includegraphics[width=\textwidth]{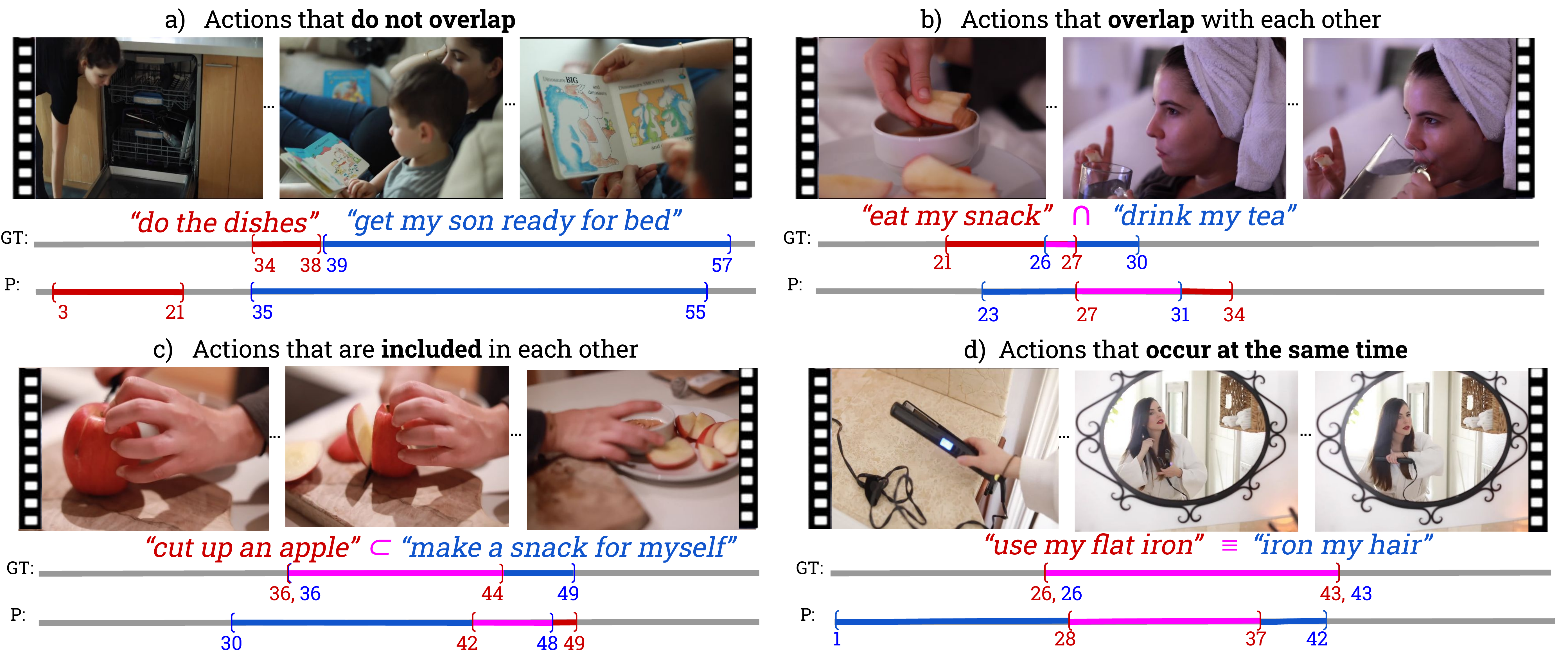}
    \caption{Randomly sampled qualitative results for different cases of action overlapping. Best viewed in color.}
    \label{fig:qual_res}
\end{figure}

\subsection{Text and Visual Features}%
\label{subsec:features}

In \cref{tab:results_weighting_feat}, we experiment with the MPU model (without applying the 2SEAL method) and look into how each modality contributes to solving this task, by removing one modality at a time from our best performing model. We also analyze other types of text embeddings. Inspired by \cite{motwani2012improving, sigurdsson2017actions}, we focus on verbs and nouns, which we extract from the actions and compute their BERT embeddings.
We observe that the visual information contributes the most to the task of action localization, as removing this information drastically lowers the model performance. Another observation is that processing the entire action is more beneficial to the model than focusing only on nouns and verbs.

\begin{table}
% \resizebox{\columnwidth}{!}{%
\setlength{\tabcolsep}{0.37em} % for the horizontal padding
{\renewcommand{\arraystretch}{1.1} % for the vertical padding
    \begin{tabular}{lrrrrr}
        & \multicolumn{5}{c}{Recall} \\
        %\cline{2-6}
        \multicolumn{1}{c}{Method} & \multicolumn{1}{c}{IoU=0.1} & \multicolumn{1}{c}{IoU=0.3} & \multicolumn{1}{c}{IoU=0.5} & \multicolumn{1}{c}{IoU=0.7} & \multicolumn{1}{c}{mIoU} \\
        \midrule
        MPU & \textbf{57.9} & \textbf{27.0} & \textbf{12.4} & \textbf{6.2} & \textbf{21.4}\\
        MPU verb only & 33.5 & 18.5 & 9.2 & 4.8 & 13.7 \\
        MPU verb+noun only & 33.8 & 18.7 & 9.8 & 4.8 & 14.0 \\
        \midrule
        MPU BERT w/o DA & 46.9 & 26.4 & 14.1 & 5.4 & 19.0 \\
        MPU ELMo & 48.5 & 23.7 & 10.6 & 6.2 & 18.4 \\
        MPU GloVe & 41.6 & 22.5 & 11.6 & 6.9 & 17.2 \\
        \midrule
        MPU video only & 41.5 & 25.4 & 13.9 & 6.8 & 18.0 \\
        MPU text only & 25.3 & 11.6 & 4.3 & 2.2 & 9.1 \\
    \end{tabular}
    }%
    % }
    \caption{Results on the test set for different variations of the input to the MPU model. ``DA'' stands for Domain Adaptation.}
    \label{tab:results_weighting_feat}
\end{table}

\subsection{Qualitative Results}%
\label{subsec:qualitative_results}

Randomly sampled results are shown in \cref{fig:qual_res}. They are grouped by the different levels of action overlapping: no overlap, intersection, inclusion and perfect overlap. From analyzing these results, a future work direction emerges: detecting which actions are likely to happen at the same time, which in turn can lead to better algorithms for action localization.

\section{Conclusion}

In this paper, we introduced a new dataset for action localization in vlogs --- a growing form of online video communication where everyday routine actions are described in language and also presented visually. %This data helps understand the relationship between what people say in the videos and what you can see in them.
Using this dataset, we addressed the task of temporal action localization in videos. % for the actions mentioned  their transcripts.
We proposed {\sc 2Seal} -- a simple yet effective method to visually localize the actions mentioned in a video transcript, which relies on both language and vision, and specifically accounts for the duration of an action for the purpose of building a more accurate system.

Through several extensive evaluations, we showed that our method improves and complements other methods by first computing the expected duration of an action, and selectively applying a language-based or multimodal model depending on the action duration. 
This work contributes to the larger body of work for multimodal understanding, and at the same time builds a large repository of vision-language representations covering a wide spectrum of actions that can be used for downstream tasks such as action recognition systems, human behavior understanding, event recognition, and others.
%sWe believe this work is a step forward for video understanding, as there are not many supervision signals available that ground video and language together, where subtitles may represent a way through however they are challenging to relate to the visual modality.
The dataset introduced in this paper, the annotation tool, and the system code are publicly available at  \url{https://github.com/MichiganNLP/vlog_action_localization}.

\section{Acknowledgments}

We thank Weiji Li for his help with the human annotations of action visibility. This research was partially supported by a grant from the Automotive Research Center (ARC) at the University of Michigan in accordance with Cooperative Agreement W56HZV-19-2-0001. 
% Identification of funding sources and other support, and thanks to
% individuals and groups that assisted in the research and the
% preparation of the work should be included in an acknowledgment
% section, which is placed just before the reference section in your
% document.

% %%
% %% The acknowledgments section is defined using the "acks" environment
% %% (and NOT an unnumbered section). This ensures the proper
% %% identification of the section in the article metadata, and the
% %% consistent spelling of the heading.
% \begin{acks}
% \end{acks}

\bibliographystyle{ACM-Reference-Format}
\bibliography{main}

\end{document}